\documentclass[12pt]{article}
\usepackage{arxiv}

\usepackage[utf8]{inputenc} %
\usepackage[T1]{fontenc}    %
\usepackage{booktabs}       %
\usepackage{amsfonts}       %
\usepackage{nicefrac}       %
\usepackage{microtype}      %

\usepackage[numbers,sort&compress]{natbib}
\usepackage{doi}
\usepackage{algorithm,algpseudocode}
\usepackage{amsmath}
\usepackage{amssymb}
\usepackage{mathtools}
\usepackage{amsthm}
\usepackage{multirow, array, threeparttable}
\usepackage[font=normalsize,skip=4pt]{caption}
\usepackage{subcaption}

\usepackage[dvipsnames]{xcolor}
\usepackage{graphicx}   %
\usepackage{hyperref}   %
\hypersetup{
    colorlinks,
    linkcolor={blue!60!black},
    citecolor={red!60!black},
    urlcolor={blue!80!black}
}
\usepackage[capitalize,nameinlink]{cleveref}

\title{Epsilon-Greedy Thompson Sampling to Bayesian Optimization}

\author{{\hspace{1mm}Bach Do}\\
	University of Houston\\
	\texttt{bdo3@uh.edu} \\
	\And
{\hspace{1mm}Taiwo Adebiyi} \\
	University of Houston\\
	\texttt{taadebiyi2@uh.edu} \\
        \And
 {\hspace{1mm}Ruda Zhang}\thanks{Corresponding author.} \\
	University of Houston\\
	\texttt{rudaz@uh.edu} \\
}

\date{}

\renewcommand{\headeright}{ }
\renewcommand{\shorttitle}{$\varepsilon$-greedy Thompson Sampling}

\begin{document}
\maketitle

\begin{abstract}
Bayesian optimization (BO) has become a powerful tool for solving simulation-based engineering optimization problems thanks to its ability to integrate physical and mathematical understandings, consider uncertainty, and address the exploitation--exploration dilemma.
Thompson sampling (TS) is a preferred solution for BO to handle the exploitation--exploration trade-off.
While it prioritizes exploration by generating and minimizing random sample paths from probabilistic models---a fundamental ingredient of BO---TS weakly manages exploitation by gathering information about the true objective function after it obtains new observations.
In this work, we improve the exploitation of TS by incorporating the $\varepsilon$-greedy policy, a well-established selection strategy in reinforcement learning.
We first delineate two extremes of TS, namely the generic TS and the sample-average TS.
The former promotes exploration, while the latter favors exploitation.
We then adopt the $\varepsilon$-greedy policy to randomly switch between these two extremes. 
Small and large values of $\varepsilon$ govern exploitation and exploration, respectively.
By minimizing two benchmark functions and solving an inverse problem of a steel cantilever beam,
we empirically show that $\varepsilon$-greedy TS equipped with an appropriate $\varepsilon$
is more robust than its two extremes,
matching or outperforming the better of the generic TS and the sample-average TS.\footnote[1]{This is the accepted version of the following article:
Do, B.; Adebiyi, T. \& Zhang, R.
Epsilon-Greedy Thompson Sampling to Bayesian Optimization.
\textit{Journal of Computing and Information Science in Engineering}, (2024),
\url{https://doi.org/10.1115/1.4066858}.
}
\end{abstract}

\keywords{Thompson sampling \and Bayesian optimization \and $\varepsilon$-greedy policy \and Exploitation--exploration dilemma \and Cyclic constitutive law }

\section{Introduction}
\label{sec:introduction}

Consider the following minimization problem:
\begin{equation}\label{eqn1}
	\begin{aligned}
		\underset{\bf x}{\min} \ \ & f({\bf x})\\
		\textrm{subject to} \ \ 
		& \bf x \in \mathcal{X}, 
	\end{aligned}
\end{equation} 
where ${\bf x} \in \mathbb{R}^d$ is the vector of $d$ input variables
selected in a bounded, compact domain $\mathcal{X} \subset \mathbb{R}^d$,
and $f({\bf x}): \mathcal{X} \mapsto \mathbb{R}$ is a real-valued objective function.
In science and engineering applications, the objective function $f({\bf x})$ is often a black-box function evaluated via costly simulations, which unfortunately hinders the use of any mature, gradient-based numerical optimization algorithms.

Bayesian optimization (BO)~\cite{Snoek2012,Shahriari2016,Frazier2018,Garnett2023,Do2023mfbo,ZhangRD2024mfml} is a global sequential optimization technique well-suited for solving small- and moderate-dimensional optimization problems with costly or black-box objective functions.
It finds application in diverse domains of science and engineering, including hyperparameter tuning of machine learning algorithms~\cite{Snoek2012}, identification of material parameters~\cite{Karandikar2022,Kuhn2022}, material design~\cite{Tran2019,Zhang2020}, airfoil design~\cite{Zheng2020}, adaptive experimental design~\cite{Greenhill2020}, and accelerator physics~\cite{Roussel2021}.
For a recent comprehensive review of its applications in engineering design, see \cite{Do2023mfbo}.
At its core, BO guides the optimization process using a probabilistic surrogate model $\widehat{f}$ of the objective function,
usually a Gaussian process (GP), coupled with an optimization policy~\cite{Garnett2023}.
Given a dataset that has several observations of the input variables and the corresponding objective function values, a GP posterior built from this dataset often serves as the probabilistic model $\widehat{f}$ encapsulating our beliefs about the black-box objective function.
The optimization policy specifying what we value in the dataset is defined through an acquisition function $\alpha(\mathbf{x})$.
This acquisition function can be deterministic or stochastic, and is cost-effective to evaluate given $\widehat{f}$, making it convenient for processing optimization.
Different considerations that should be taken into account when formulating $\alpha(\mathbf{x})$ from $\widehat{f}$ include, for example, the value of the objective function~\cite{Hennig2022}, the information about the minimum location~\cite{Villemonteix2009,Hennig2012,HernandezLobato2014}, and the information about the minimum value~\cite{WangZ2017}.

Various algorithms have been developed to address the classic exploitation--exploration dilemma in BO, where exploitation involves selecting new solutions anticipated to improve the objective function value immediately and exploration focuses on sampling new solutions to reduce uncertainty in predictions of the objective function for a long-term improvement in the solution.
An effective BO algorithm is deemed to balance these opposing concerns~\cite{Garnett2023}.    
Notable deterministic acquisition functions, such as expected improvement (EI)~\cite{Jones1998}, weighted EI~\cite{Sobester2005}, GP upper confidence bound (GP-UCB)~\cite{Srinivas2010}, knowledge gradient (KG)~\cite{Frazier2008}, and likelihood-weighting~\cite{Blanchard2021}, can manage the exploitation--exploration trade-off.
However, they are considered myopic policies, and therefore tend to favor exploitation~\cite{Hennig2022}.

Thompson sampling (TS) is a preferred algorithm to address the exploitation--exploration dilemma in solving multi-armed bandit problems~\cite{Thompson1933,Chapelle2011,Russo2014,Russo2018}.
More specifically, TS selects arms (or actions) from  
a finite set of possible arms over a limited number of iterations.
Each arm corresponds to a stochastic reward drawn from an unknown distribution.
The goal is to craft a sequence of arms that maximizes the cumulative reward assuming that the rewards are independent of time and conditionally independent given the selected arms.
When applied to GP-based BO, TS (or GP-TS \cite{Garnett2023}) generates a sequence of new solution points using the mechanism that involves random sampling from unknown posterior distributions of the global minimum location $\bf{x}^\star$~\cite{Kandasamy2018}, which is due to the imperfect knowledge of $f$ described by the probabilistic model $\widehat{f}$.
Rather than finding the posterior distributions directly, TS generates functions from $\widehat{f}$ and minimizes them for samples of $\bf{x}^\star$.
Thus, we can consider TS a BO method that minimizes stochastic acquisition functions generated from the model posteriors for selecting new solution points.
It is also worth noting that several information-theoretic optimization policies of BO, such as predictive entropy~\cite{HernandezLobato2014} and max-value entropy~\cite{WangZ2017}, compute their acquisition functions based on a set of samples generated by TS.

While TS naturally manages the exploitation--exploration trade-off,
its randomness can reduce the role of exploitation in optimization.
When the GP posterior exhibits high uncertainty at the beginning of the optimization process, TS prioritizes exploration as it can diversify the selection of new solutions when limited information has been gained about the optimal solution $\bf{x}^\star$.
As the number of observations increases and the GP posterior becomes concentrated, the algorithm transitions to exploiting knowledge about the true objective function.
This exploitation strategy, however, is inferior due to the randomness of TS, motivating the quest for an intervention to improve its exploitation. 

In this work, we incorporate the $\varepsilon$-greedy policy into TS to improve its exploitation.
This policy is a selection strategy of reinforcement learning to address the tension between exploitation and exploration~\cite{Sutton2018}.
Given $\varepsilon \in (0,1)$, the policy chooses an action by either maximizing an average reward function with probability $1-\varepsilon$ (i.e., exploitation or greedy) or selecting it randomly with probability $\varepsilon$ (i.e., exploration).
The selection strategy is pure exploitation when $\varepsilon = 0$, or pure exploration when $\varepsilon = 1$.  
Similarly, our proposed approach implements the generic TS (with probability $\varepsilon$) for exploration and a new fashion of TS called sample-average TS (with probability $1-\varepsilon$) for exploitation.

Several works have explored the $\varepsilon$-greedy policy in BO and multi-armed bandit problems. 
De Ath et al.~\cite{DeAth2021} proposed two schemes for applying the policy to BO. 
The first scheme performs exploration (with probability $\varepsilon$) by randomly selecting a point on the Pareto frontier obtained by simultaneously minimizing the posterior mean and maximizing the posterior standard deviation.
The second scheme performs exploration by randomly selecting a point in the input variable space.
Both schemes implement exploitation (with probability $1-\varepsilon$) by minimizing the posterior mean function.
Jin et al.~\cite{Jin2023} introduced the so-called $\varepsilon$-exploring TS ($\varepsilon$-TS) to multi-armed bandit problems.
Given the posterior distributions of arms, exploration (with probability $\varepsilon$) sets the estimated reward function for each arm as a random sample drawn from the associated posterior distribution, while exploitation (with probability $1-\varepsilon$) sets the estimated reward function as the sample mean function. 
Yet the performance of $\varepsilon$-greedy TS for BO remains unexplored.

Our contributions are as follows:
(1) $\varepsilon$-greedy TS to BO, which is a simple, effective method to improve the exploitation of TS;
and (2) empirical evaluations demonstrating that $\varepsilon$-greedy TS with an appropriate $\varepsilon$
is more robust than its two extremes (i.e., the generic TS and the sample-average TS),
matching or outperforming the better of the two across various problems.

The rest of this paper progresses as follows.
\cref{sec:background} provides a background essential for the development of $\varepsilon$-greedy TS.
\cref{sec:method} describes the method in detail.
\cref{sec:experiments} presents the empirical performance of $\varepsilon$-greedy TS on minimizing two benchmark functions and finding constitutive parameters for a steel cantilever beam.
Finally, \cref{sec:conclusions} concludes this paper.

\section{Background}
\label{sec:background}
This section provides an overview of GP modeling (\cref{sec:GP}), the implementation of the generic TS for BO (\cref{sec:TS}), and a simple method to generate sample paths from GP posteriors for use of TS (\cref{sec:sampling}).

\subsection{Overview of Gaussian Processes}
\label{sec:GP}

Let $\mathcal{D}=\left\{ \left( {\bf X},{\bf y}\right) \right\}=\left\{ \left({\bf x}^i,y^i\right) \right\}_{i=1}^N$ denote a training dataset, where $\textbf{x}^i$ are observations of a $d$-dimensional vector of input variables and $y^i$ the corresponding observations of the objective function.
We wish to build from $\mathcal{D}$ an observation model $y({\bf x})=f({\bf x}) + \varepsilon_\text{n}: \mathbb{R}^d \mapsto \mathbb{R}$, where $\varepsilon_\text{n} \sim \mathcal{N}(0,\sigma^2_\text{n})$ is additive zero-mean Gaussian noise with variance $\sigma^2_\text{n}$.
This observation noise is assumed to be independent and identically distributed.

A GP model assumes that any finite subset of an infinite set of objective function values has a joint Gaussian distribution~\cite{Rasmussen2006}.
Oftentimes this assumption is encoded using the following GP prior:
\begin{equation}\label{eqn2}
	f(\cdot) \sim \mathcal{GP} \left(0,\kappa(\cdot,\cdot|\boldsymbol{\theta})\right),
\end{equation}
where $\kappa({\bf x},{\bf x}'|\boldsymbol{\theta}) = \text{cov}[f({\bf x}),f({\bf x}')]: \mathbb{R}^d \times \mathbb{R}^d \mapsto \mathbb{R} $ is a positive semi-definite covariance function parameterized by a vector $\boldsymbol{\theta}$ of hyperparameters.
Common-used covariance functions include squared exponential (SE), Matérn 3/2, Matérn 5/2, and automatic relevance determination squared exponential (ARD SE)~\cite{Garnett2023,Rasmussen2006}.
By conditioning on $\mathcal{D}$ the model prior given in \cref{eqn2} and utilizing the observation noise assumption, it is straightforward to show that the vector of observations $\left[y({\bf x}^1),\dots,y({\bf x}^N)\right]^\intercal$ is distributed according to an $N$-variate Gaussian with zero mean and covariance matrix ${\bf C} = {\bf K} + \sigma^2_\text{n} {\bf I}_N$, where the $(i,j)$th element of ${\bf K}$ is $\kappa({\bf x}^i,{\bf x}^j|\boldsymbol{\theta})$ and ${\bf I}_N$ the $N$-by-$N$ identity matrix.
By further applying the conditional multivariate Gaussian~\cite{Bishop2006}, we obtain the posterior predictive distribution of the objective function value at an unseen input variable vector ${\bf x}_\star$, i.e., $p\left(f({\bf x}_\star)|{\bf x}_\star,\mathcal{D}\right) = \mathcal{N}\left(\mu_\text{f}({\bf x}_\star),\sigma_\text{f}^2({\bf x}_\star)\right)$, which encodes information about the model we wish to build. The mean and variance of the predictive distribution read 
\begin{equation}\label{eqn3}
	\mu_\text{f}({\bf x}_\star)= {\bf K_\star}^\intercal {\bf C}^{-1} {\bf Y},
\end{equation}
\begin{equation}\label{eqn4}
	\sigma_\text{f}^2({\bf x}_\star)=\kappa({\bf x}_\star,{\bf x}_\star|\boldsymbol{\theta})
	-{\bf K_\star}^\intercal {\bf C}^{-1} {\bf K_\star},
\end{equation}
where
\begin{equation}\label{eqn5}
	{\bf K_\star} = \left[\kappa({\bf x}_\star,{\bf x}^1|\boldsymbol{\theta}),\dots,\kappa({\bf x}_\star,{\bf x}^N|\boldsymbol{\theta})\right]^\intercal.
\end{equation}

Training GP models involves finding an optimal set of hyperparameters $\boldsymbol{\theta}$ that ensures robust predictive performance of trained GPs across unseen input variables.
Exact training methods emulating all possible hyperparameter hypotheses are often impractical, leading to the prevalent use of approximate training methods.
These approximate training methods fall into two main categories: deterministic approximations and Monte Carlo methods~\cite{Mackay2003}.
Deterministic approximations include the maximum likelihood method identifying a set of hyperparameters that maximizes the parameter likelihood $p(\mathcal{D}|\boldsymbol{\theta})$, and Laplace's method approximating the posterior $p(\boldsymbol{\theta}|\mathcal{D})$ as a multivariate Gaussian with a posterior mean equal to the maximum likelihood estimation and a posterior covariance estimated from Laplace approximation.
Monte-Carlo methods, such as importance sampling, the Metropolis, Gibbs sampling, and slice sampling, generate approximate samples of $\boldsymbol{\theta}$ from the posterior $p(\boldsymbol{\theta}|\mathcal{D})$ given the likelihood $p(\mathcal{D}|\boldsymbol{\theta})$ and a prior $p(\boldsymbol{\theta})$.
Several of the aforementioned approximate training methods have been incorporated in reliable GP toolboxes such as DACE~\cite{Lophaven2002}, GPML~\cite{Rasmussen2010}, GPstuff~\cite{Vanhatalo2013}, pyGPs~\cite{Neumann2015}, and GPflow~\cite{Matthews2017}.

\begin{figure}[t]
	\centering
	\begin{subfigure}[c]{0.49\textwidth}
		\centering
		\includegraphics[width=\hsize]{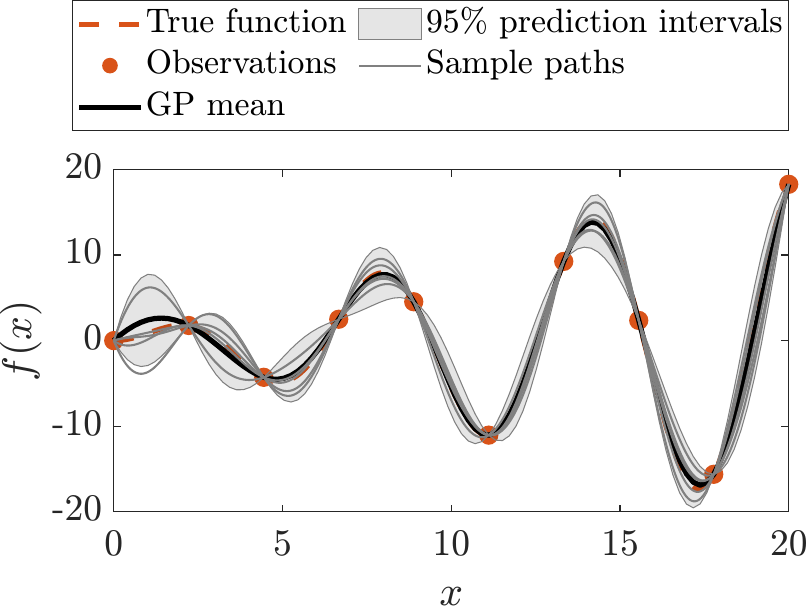}
		\caption{}
		\label{fig:paths_a}
	\end{subfigure}
	\begin{subfigure}[c]{0.49\textwidth}
		\centering
		\includegraphics[width=\hsize]{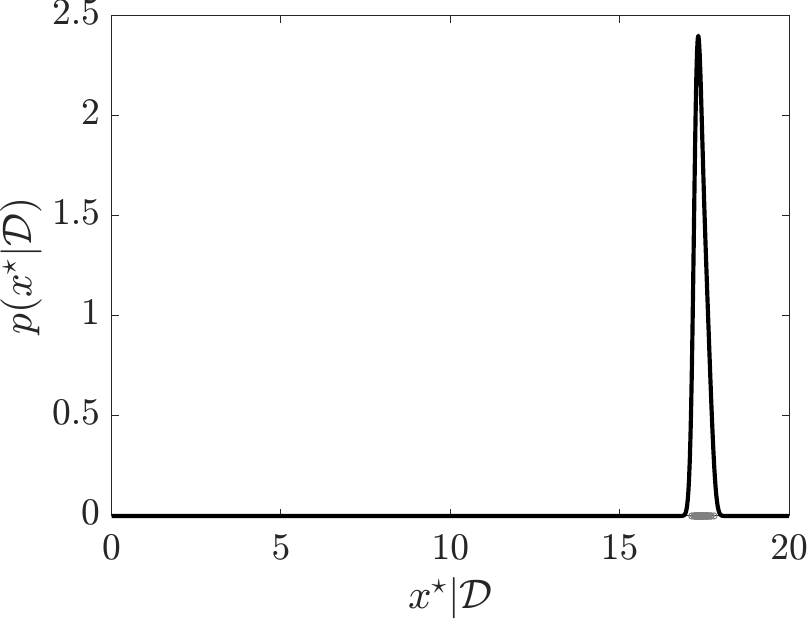}
		\caption{}
		\label{fig:paths_b}
	\end{subfigure}
	\caption{Sample paths from the GP posterior for $f(x) = x\sin(x)$ and distribution of their minimum locations. (a) GP predictions and five sample paths drawn from the GP posterior; (b) Approximate conditional distribution $p( x^\star|\mathcal{D})$ obtained from minimum locations of 50 sample paths.} 
	\label{fig:paths}
\end{figure}

\begin{algorithm}[t]
	\caption{Generic Thompson sampling for Bayesian optimization}
	\label{alg:TS}
	\begin{algorithmic}[1]
		
		\State \textbf{input:} input variable domain $\mathcal{X}$, number of initial observations $N$, threshold for number of BO iterations $K$, standard deviation of observation noise $\sigma_\text{n}$
		
		\State Generate $N$ initial samples of ${\bf x}$
		
		\For {$i=1:N$} 
		\State $y^i \gets f({\bf x}^i) + \varepsilon_\text{n}^i$ 
		\EndFor
		
		\State $\mathcal{D}^0 \gets \{{\bf x}^i,y^i\}_{i=1}^N$
		
		\State $\{{\bf x}_{\min},y_{\min}\} \gets \min\{y^i,\, i=1,\dots, N\}$
		
		\For {$k=1:K$} 
		\State Build a GP posterior $\widehat{f}^k({\bf x})|\mathcal{D}^{k-1}$
		\State Generate a sample path $g({\bf x}|\mathcal{D}^{k-1})$ from $\widehat{f}^k({\bf x})|\mathcal{D}^{k-1}$ \label{alg:TS_L10}
		\State ${\bf x}^{k} \gets \underset{{\bf x}}{\mathrm{arg\,min}} \ \ g({\bf x}|\mathcal{D}^{k-1})$ s.t. ${\bf x} \in \mathcal{X}$; ${\bf x} \notin \mathcal{D}^{k-1}$
		\State $y^{k} \gets f({\bf x}^{k}) + \varepsilon_\text{n}^k$
		\State $\mathcal{D}^k\gets\mathcal{D}^{k-1} \cup \{{\bf x}^{k},y^{k}\}$
		\State $\{{\bf x}_{\min},y_{\min}\} \gets \min\{y_{\min},y^{k}\}$
		\EndFor
		
		\State \textbf{return} $\{{\bf x}_{\min},y_{\min}\}$
	\end{algorithmic}
\end{algorithm}
\subsection{Bayesian Optimization via Thompson Sampling}
\label{sec:TS}

As briefly described in \cref{sec:introduction}, the generic TS generates a sequence of solution points ${\bf x}^k$ $(k=1,\dots,K)$ by randomly sampling from an unknown posterior distribution $p({\bf x}^\star|\mathcal{D}^{k-1})$ of the global minimum ${\bf x}^\star$, where $K$ represents a finite budget on the number of BO iterations.
Leveraging the fact that ${\bf x}^\star$ is fully determined by the objective function, the generic TS follows two simple steps in each iteration.
It first generates a sample path from the GP posterior $\widehat{f}^k({\bf x})|\mathcal{D}^{k-1}$ for which the detailed implementation is described in \cref{sec:sampling}.
It then minimizes the generated sample path to find a minimum location ${\bf x}^\star$.
By doing so, ${\bf x}^\star$, assigned as the new solution ${\bf x}^k$, is a sample from $p({\bf x}^\star|\mathcal{D}^{k-1})$.
\cref{alg:TS} summarizes the implementation of the generic TS.

\Cref{fig:paths} shows sample paths generated from the GP posterior of univariate function $f(x)=x\sin{x}$ for $x \in [0,20]$ and an approximate distribution derived from their minimum locations.
Specifically, we build from ten observations a GP model with zero mean and SE covariance function.
From this GP model, we generate 50 sample paths, and five of them are shown in \cref{fig:paths_a}. 
Minimizing the generated sample paths and utilizing a kernel density estimation on the obtained solutions result in an approximate distribution $p({\bf x}^\star|\mathcal{D})$, as shown in \cref{fig:paths_b}.
Note that the generic TS in \cref{alg:TS} generates and minimizes only one sample path in each iteration and does not attempt to approximate the posterior distribution of the minimum location. The following section describes how we can randomly generate sample path $g({\bf x}|\mathcal{D}^{k-1})$ in \cref{alg:TS_L10} of \cref{alg:TS}.

\subsection{Sampling from Gaussian Process Posteriors}
\label{sec:sampling}

\begin{table*}[t]
	\caption{Spectral density functions for common-used stationary covariance functions~\cite{RiutortMayol2022}.}
	\label{table1}
	\centering
	\begin{threeparttable}
	\renewcommand{\arraystretch}{1}
	\begin{tabular}{lll}
		\hline\noalign{\smallskip}
		 {Covariance} & $\kappa({\bf x},{\bf x}'|\boldsymbol{\theta})$ & Spectral density function $S({\bf s})$\\
		\hline\noalign{\smallskip}
		
		SE & $\sigma_\text{f}^2 \exp{\left(-\frac{1}{2}\frac{r^2}{l^2}\right)}$ \tnote{(1)}   &$\sigma_\text{f}^2(\sqrt{2\pi})^d l^d \exp{\left(-\frac{1}{2} l^2 {\bf s}^\intercal {\bf s}\right)}$  \\
		\noalign{\smallskip}
		
		ARD SE & $\sigma_\text{f}^2 \exp{\left(-\frac{1}{2}\sum_{i=1}^{d}\frac{(x_i-x_i')^2}{l_i^2}\right)}$ \tnote{(2)}    &$\sigma_\text{f}^2 (\sqrt{2\pi})^d \left(\prod_{i=1}^{d}l_i\right) \exp{\left(-\frac{1}{2} \sum_{i=1}^{d}l_i^2 s_i^2\right)}$  \\
		\noalign{\smallskip}
		
		Matérn 3/2 & $\sigma_\text{f}^2 \left( 1+ \frac{\sqrt{3}r}{l}\right) \exp{\left(-\frac{\sqrt{3}r}{l}\right)}$ \tnote{(3)} & $\sigma_\text{f}^2 \frac{2^d \pi^{d/2} \Gamma\left(\frac{d+3}{2}\right)3^{3/2}}{\frac{1}{2} \sqrt{\pi} l^3} \left(\frac{3}{l^2} + {\bf s}^\intercal {\bf s} \right)^{-\frac{d+3}{2}}$ \tnote{(4)} \\
		\noalign{\smallskip}
		
		Matérn 5/2 & $\sigma_\text{f}^2 \left( 1+ \frac{\sqrt{5}r}{l} + \frac{5 r^2}{3 l^2}\right) \exp{\left(-\frac{\sqrt{5}r}{l}\right)}$ \tnote{(3)} & $\sigma_\text{f}^2 \frac{2^d \pi^{d/2} \Gamma\left(\frac{d+5}{2}\right)5^{5/2}}{\frac{3}{4} \sqrt{\pi} l^5} \left(\frac{5}{l^2} + {\bf s}^\intercal {\bf s} \right)^{-\frac{d+5}{2}}$ \tnote{(4)}\\
		\hline\noalign{\smallskip}
		
	\end{tabular}
	\begin{tablenotes}
		\item[(1)] $r = \sqrt{({\bf x}-{\bf x}')^\intercal ({\bf x}-{\bf x}')}$ and $\boldsymbol{\theta} = [\sigma_\text{f},l]^\intercal$, where $l>0$ and $\sigma_\text{f}>0$  are length scale and marginal standard deviation, respectively
		
		\item[(2)] $\boldsymbol{\theta} = [\sigma_\text{f},l_1,\dots,l_d]^\intercal$, where $l_i>0$ $(i=1,\dots,d)$ is length scale for the $i$th input variable
		
		\item[(3)]	$r = \sqrt{({\bf x}-{\bf x}')^\intercal ({\bf x}-{\bf x}')}$ and $\boldsymbol{\theta} = [\sigma_\text{f},l]^\intercal$
		
		\item[(4)] $\Gamma(\cdot) = $ Gamma function
	\end{tablenotes}
\end{threeparttable}
\end{table*}

Given the GP posterior $\widehat{f}^k({\bf x})|\mathcal{D}^{k-1}$ characterized by a stationary covariance function, we follow the spectral sampling approach to generate a sample path $g({\bf x}|\mathcal{D}^{k-1})$, see e.g., \cite{Rahimi2007} and \cite{HernandezLobato2014}.
This approach approximates the GP prior in \cref{eqn2} using a Bayesian linear model of randomized basis functions to avoid the computational cost due to exhaustive sampling from marginal distributions of the objective function values at finite sets of input locations, which scales cubically in the number of input locations.
Such an approximation is rooted in Bochner’s theorem that guarantees the existence of a Fourier dual $S({\bf s})$ (${\bf s} \in \mathbb{R}^d$) of the stationary covariance function, which is called spectral density when a finite non-negative Borel measure is interpreted as a distribution~\cite{Wendland2004}.
The spectral density functions associated with several covariance functions of Matérn class and ARD SE are listed in \cref{table1}.
Once $S({\bf s})$ is determined, we can represent $\kappa({\bf x},{\bf x}'|\boldsymbol{\theta})$ using the following randomized feature map~\cite{Rahimi2007}:
\begin{equation}\label{eqn6}
	\kappa({\bf x},{\bf x}'|\boldsymbol{\theta}) = \boldsymbol{\phi}({\bf x})^\intercal \boldsymbol{\phi}({\bf x}'),
\end{equation}
where the feature map $\boldsymbol{\phi}({\bf x})\in \mathbb{R}^{N_\text{p}}$ can be approximated by~\cite{Rahimi2007,HernandezLobato2014}
\begin{equation}\label{eqn7}
	\boldsymbol{\phi}({\bf x}) = \sqrt{2 \kappa(0|\boldsymbol{\theta})/N_\text{p}} \cos \left({\bf W} {\bf x} + {\bf b}\right).
\end{equation}
Here ${\bf W} \in \mathbb{R}^{N_\text{p} \times d}$ and ${\bf b}\in \mathbb{R}^{N_\text{p}}$ stack $N_\text{p}$ spectral points generated from the normalized spectral density $p({\bf s})=S({\bf s})/\kappa(0|\boldsymbol{\theta})$ and $N_\text{p}$ points drawn from the uniform distribution $\mathcal{U}[0,2\pi]$, respectively. $\kappa(0|\boldsymbol{\theta})$ is well defined because $\kappa({\bf x},{\bf x}'|\boldsymbol{\theta})$ is a stationary covariance function satisfying $\kappa({\bf x},{\bf x}'|\boldsymbol{\theta}) = \kappa(\left\|{\bf x}- {\bf x}' \right\||\boldsymbol{\theta})$.

\begin{figure}[t!]
	\centering
	\begin{subfigure}[c]{0.49\textwidth}
		\centering
		\includegraphics[width=\hsize]{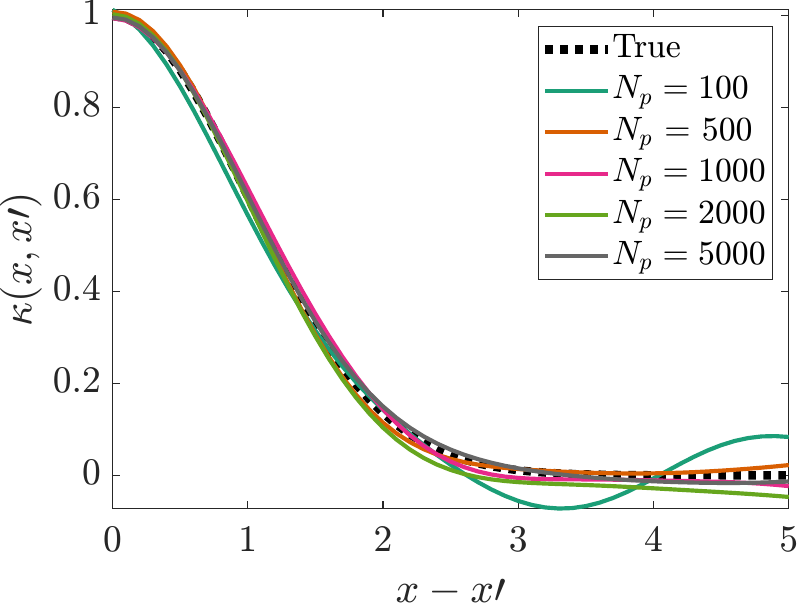}
		\caption{}
		\label{fig:approximatekernel_a}
	\end{subfigure}
	\centering
	\begin{subfigure}[c]{0.49\textwidth}
		\centering
		\includegraphics[width=\hsize]{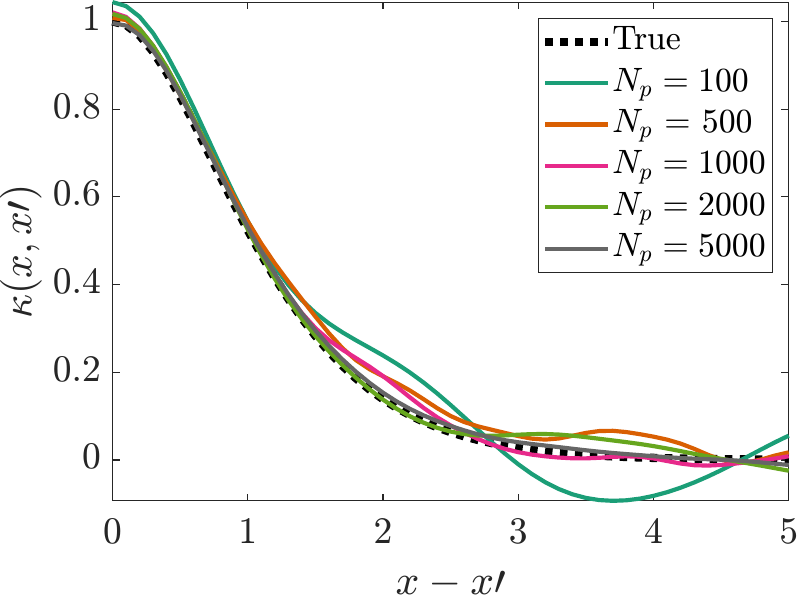}
		\caption{}
		\label{fig:approximatekernel_b}
	\end{subfigure}
	\caption{Approximate covariance functions using random features from different numbers of spectral point samples. (a) SE; (b) Matérn 5/2.}
	\label{fig:approximatekernel}
\end{figure}

\begin{algorithm}[t]
	\caption{Generation of sample paths using random features}
	\label{alg:spectral}
	\begin{algorithmic}[1]
		\State \textbf{Input:} dataset $\mathcal{D}^{k-1}$, type of stationary covariance function $\kappa(\cdot,\cdot|\boldsymbol{\theta})$, number of spectral points $N_\text{p}$, standard deviation of observation noise $\sigma_\text{n}$
		
		\State Build a GP model from $\mathcal{D}^{k-1}$
		\State Formulate $p(\bf s)$ from the posterior covariance function $\kappa(\cdot,\cdot|\boldsymbol{\theta})$, see \cref{table1}
		
		\For {$j=1:N_\text{p}$} 
		\State $[{\bf W}]_j \sim p(\bf s)$
		\State $[{\bf b}]_j \sim \mathcal{U}[0,2\pi]$
		\EndFor
		
		\State Formulate $\boldsymbol{\phi}({\bf x})$, see \cref{eqn7}
		
		\State $\boldsymbol{\Phi} \gets [\boldsymbol{\phi}({\bf x}),\dots,\boldsymbol{\phi}({\bf x}^N)]^\intercal$
		
		\State Compute $\boldsymbol{\mu}_{\beta}$ and $\boldsymbol{\Sigma}_{\beta}$, see \cref{eqn9} \label{alg:spectral_L10}
		
		\State $\boldsymbol{\beta} \sim \mathcal{N}(\boldsymbol{\mu}_{\beta}, \boldsymbol{\Sigma}_{\beta})$
		
		\State \textbf{return} $g({\bf x})|\mathcal{D}^{k-1} \gets \boldsymbol{\beta}^\intercal \boldsymbol{\phi}({\bf x})$
	\end{algorithmic}
\end{algorithm}

\Cref{fig:approximatekernel} shows approximate SE and Matérn 5/2 covariance functions computed from \cref{eqn6} for different numbers of spectral point samples.
We see that the approximate covariance functions converge to the true ones when the number of spectral points increases.

Once the approximate feature map is formulated, the GP prior as a kernel machine can be approximated by the following Bayesian linear model: 
\begin{equation}\label{eqn8}
	f({\bf x}) \approx \boldsymbol{\beta}^\intercal \boldsymbol{\phi}({\bf x}),
\end{equation}
where the prior of $\boldsymbol{\beta}$ is $\mathcal{N}({\bf 0}, {\bf I}_{N_\text{p}})$.
By further conditioning \cref{eqn8} on the data, we obtain the following mean and covariance of the posterior of $\boldsymbol{\beta}$~\cite{HernandezLobato2014}
\begin{subequations}\label{eqn9}
	\begin{align}
		\boldsymbol{\mu}_{\beta} & = \left(\boldsymbol{\Phi}^\intercal \boldsymbol{\Phi} +\sigma_\text{n}^2 {\bf I}_{N_\text{p}}\right)^{-1} \boldsymbol{\Phi}^\intercal {\bf y},\\
		\boldsymbol{\Sigma}_{\beta} & = \left(\boldsymbol{\Phi}^\intercal \boldsymbol{\Phi} +\sigma_\text{n}^2 {\bf I}_{N_\text{p}}\right)^{-1} \sigma_\text{n}^2,
	\end{align}
\end{subequations}
where $\boldsymbol{\Phi} = [\boldsymbol{\phi}({\bf x}),\dots,\boldsymbol{\phi}({\bf x}^N)]^\intercal \in \mathbb{R}^{N \times N_\text{p}}$. To this end, we can approximate the GP posterior using a Bayesian linear model that weights the randomized basis functions using samples from the posterior of $\boldsymbol{\beta}$.

The generation of $g({\bf x}|\mathcal{D}^{k-1})$ using random features is detailed in \cref{alg:spectral}.
Note that \cref{alg:spectral_L10} of \cref{alg:spectral} involves the Cholesky decomposition of matrix $\boldsymbol{\Phi}^\intercal \boldsymbol{\Phi} +\sigma_\text{n}^2 {\bf I}_{N_\text{p}}$ to facilitate the calculation of its inverse. 
Alternative posterior sampling techniques such as the decoupled method~\cite{Wilson2020}, which generates GP posteriors by updating the corresponding GP prior samples, can be adopted to increase the sample path accuracy.

\section{Epsilon-greedy Thompson Sampling}
\label{sec:method}

In this section, we first introduce another version of TS called sample-average TS (\cref{sec:method-avaraging}). We then describe $\varepsilon$-greedy TS using the sample-average and generic TS in the context of $\varepsilon$-greedy policy (\cref{sec:method-epsgreedyTS}).

\subsection{Sample-average Thompson Sampling}
\label{sec:method-avaraging}

To enhance the exploitation of TS, we implement another version of TS called sample-average TS (or averaging TS)~\cite{Balandat2020}.
Specifically, we call \cref{alg:spectral} to generate in \cref{alg:TS_L10} of \cref{alg:TS} a total of $N_\text{s}$ sample paths $h^s({\bf x}|\mathcal{D}^{k-1})$ ($s=1,\dots,N_\text{s}$). These sample paths are used to define the following average sample path:
\begin{equation}\label{eqn10}
	g({\bf x}|\mathcal{D}^{k-1}) = \frac{1}{N_\text{s}}\sum_{s=1}^{N_\text{s}} h^s({\bf x}|\mathcal{D}^{k-1}).
\end{equation}
Another approach to generate an average sample path is to use a decoupled representation via pathwise conditioning \cite{Adebiyi2024bdu,Adebiyi2024roots},
which can be computed at the same cost of generating one sample path,
regardless of the nominal number $N_s$ of sample paths being averaged.

We then minimize the average sample path $g({\bf x}|\mathcal{D}^{k-1})$ to find a new solution in each iteration of BO.
This approach differs from that of Balandat et al.~\cite{Balandat2020} which generates sample paths using randomized quasi Monte-Carlo techniques.

The sample-average TS and the generic TS can be considered two extremes of TS. If $N_\text{s} = \infty$, $g({\bf x}|\mathcal{D}^{k-1})$ in \cref{eqn10} is indeed the GP posterior mean in \cref{eqn3} whose minimum promotes exploitation.
If $N_s = 1$, the sample-average TS recovers the generic TS that favors exploration. 
Thus, there exists an optimal state between these two extremes that corresponds to an unknown value of $N_s$ where TS balances exploitation and exploration.
However, we do not attempt to find such an optimal state in this work.
We instead set $N_\text{s}$ at a sufficiently large value, say $N_\text{s}=50$, to enforce the exploitation of TS.

\subsection{$\varepsilon$-greedy Thompson Sampling}
\label{sec:method-epsgreedyTS}

The two distinct extremes of TS motivate the use of $\varepsilon$-greedy policy to randomly switch between them.
Specifically, we implement the generic TS with probability $\varepsilon$ to explore the input variable space.
We invoke the sample-average TS with probability $1-\varepsilon$ to guide the search toward exploitation.
We do not use the posterior mean in \cref{eqn3} for exploitation because we observe that computing its derivatives is more expensive than computing the derivatives of the average sample path in \cref{eqn10}, which adds computational cost if optimization is performed via a gradient-based algorithm.
It is also straightforward to recover the generic TS from $\varepsilon$-greedy TS by simply setting $\varepsilon=1$ or $N_\text{s}=1$.
\cref{alg:epsilon_greedyTS} details the proposed $\varepsilon$-greedy TS.

The proposed algorithm addresses the exploitation--exploration dilemma by the following factors: (1) $\varepsilon$ \textendash a small value of $\varepsilon$ promotes exploitation and (2) $N_\text{s}$ \textendash a sufficiently large value of $N_\text{s}$ encourages exploitation.

\begin{algorithm}[t]
	\caption{$\varepsilon$-greedy Thompson sampling for Bayesian optimization}
	\label{alg:epsilon_greedyTS}
	\begin{algorithmic}[1]
		
		\State \textbf{input:} input variable domain $\mathcal{X}$, number of initial observations $N$, threshold for number of BO iterations $K$, number of spectral points $N_\text{p}$, number of sample paths $N_\text{s}$, value of $\varepsilon \in (0,1)$, standard deviation of observation noise $\sigma_\text{n}$
		
		\State Generate $N$ samples of ${\bf x}$
		
		\For {$i=1:N$} 
		\State $y^i \gets f({\bf x}^i) + \varepsilon_\text{n}^i$ 
		\EndFor
		
		\State $\mathcal{D}^0 \gets \{{\bf x}^i,y^i\}_{i=1}^N$
		
		\State $\{{\bf x}_{\min},y_{\min}\} \gets \min\{y^i,\, i=1,\dots, N\}$
		
		\For {$k=1:K$} 
		\State Build a GP posterior $\widehat{f}^k({\bf x})|\mathcal{D}^{k-1}$
		\State Generate $r \sim \mathcal{U}[0,1]$
		\If {$r \leq \varepsilon$}
		\State Sample $g({\bf x}|\mathcal{D}^{k-1})$
		\Else
		\For {$s=1:N_\text{s}$} 
		\State Sample $h^s({\bf x}|\mathcal{D}^{k-1})$
		\EndFor
		\State $g({\bf x}|\mathcal{D}^{k-1}) \gets \frac{1}{N_\text{s}}\sum_{s=1}^{N_\text{s}} h^s({\bf x}|\mathcal{D}^{k-1})$
		\EndIf
		\State ${\bf x}^{k} \gets \underset{{\bf x}}{\mathrm{arg\,min}} \ \ g({\bf x}|\mathcal{D}^{k-1})$ s.t. ${\bf x} \in \mathcal{X}$; ${\bf x} \notin \mathcal{D}^{k-1}$
		\State $y^{k} \gets f({\bf x}^{k}) + \varepsilon_\text{n}^k$
		\State $\mathcal{D}^k\gets\mathcal{D}^{k-1} \cup \{{\bf x}^{k},y^{k}\}$
		\State $\{{\bf x}_{\min},y_{\min}\} \gets \min\{y_{\min},y^{k}\}$
		\EndFor
		
		\State \textbf{return} $\{{\bf x}_{\min},y_{\min}\}$
	\end{algorithmic}
\end{algorithm}

\section{Numerical Examples}
\label{sec:experiments}

We test the empirical performance of $\varepsilon$-greedy TS on two challenging minimization problems of the 2d Ackley and 6d Rosenbrock functions (\cref{sec:example-functions}), and on the identification problem of a steel cantilever beam subjected to cyclic loading (\cref{sec:example-beam}).

\subsection{Benchmark Functions}
\label{sec:example-functions}

Consider minimization problems of the 2d Ackley and 6d Rosenbrock functions~\cite{Surjanovic2013}.
The analytical expressions for these functions and their global minimums are given as follows: 

\textbf{2d Ackley function}
\begin{equation}\label{eqn11}
	\begin{aligned}
		f({\bf x}) = & -a \exp\left(-b\sqrt{\frac{1}{2} \sum_{i=1}^{2}x^2_i}\right) \\
		&- \exp\left(\frac{1}{2} \sum_{i=1}^{2}\cos(cx_i)\right) +a + \exp(1),
	\end{aligned}
\end{equation}
where $a = 20$, $b = 0.2$, and $c = 2\pi$.
The function is evaluated on $\mathcal{X}=[-10,10]^2$ and has a global minimum at ${\bf x}^\star = [0,0]^\intercal$ with $f^\star = f({\bf x}^\star) = 0$.

\textbf{6d Rosenbrock function}
\begin{equation}\label{eqn12}
	f({\bf x}) = \sum_{i=1}^{5} \left[100(x_{i+1}-x_i^2)^2 + (x_i - 1)^2\right],
\end{equation}
where $\mathcal{X}=[-5,10]^6$. The function has a global minimum at ${\bf x}^\star = [1,\dots,1]^\intercal$ with $f^\star = f({\bf x}^\star) = 0$.

While $\varepsilon$-greedy TS can be applied to any stationary covariance function listed in \cref{table1}, we restrict our experiments to the ARD SE covariance function, assuming no specific prior knowledge about the characteristics of the objective functions.
We call the DACE toolbox~\cite{Lophaven2002} to find optimal GP hyperparameters for each minimization problem.
We set $\sigma_\text{n} = 10^{-3}$ for $z-$score standardized output observations and generate $10^3$ spectral points to approximate the sample paths of GP posteriors. 
To minimize the generated sample paths, we use the DIRECT algorithm~\cite{Finkel2006} whose output solution is set as the initial point of an interior-point optimizer.
Parameters for the DIRECT algorithm including the function tolerance, maximum number of function evaluations, maximum number of iterations, and maximum number of rectangle divisions are $10^{-9}$, $10^3d$, $10^4d$, and $10^4d$, respectively.
Parameters for the interior-point optimizer including the function tolerance, optimality tolerance, maximum number of iterations, maximum number of function evaluations, and termination tolerance on ${\bf x}$ are $10^{-12}$, $10^{-12}$, $200$, $5\times10^{3}$, and $10^{-12}$, respectively.  
We carry out all experiments for the benchmark functions using the Carya Cluster of the University of Houston.
We compare the optimization results obtained from $\varepsilon$-greedy TS with those from other BO methods, including EI, the lower confidence bound (LCB), the averaging TS, and the generic TS.

To determine an appropriate value of $N_\text{s}$ in \cref{eqn10} which trade-offs between the exploitation and computational cost, our preliminary experiments investigate how changes in $N_\text{s}$ affect the optimization results for a fixed value of $\varepsilon$.
We observe that when the number of sample paths reaches a sufficiently large value of $N_\text{s}=50$, the solutions exhibit less sensitivity to changes in $N_\text{s}$.

To ensure fair comparisons, we randomly generate 100 initial datasets for each problem using Latin hypercube sampling~\cite{Forrester2008}, and start each BO method from each of these datasets.
In each BO iteration, we record the best-found value of observation error $\log(y_{\min}-f^\star)$ and the corresponding solution vector. Here $y_{\min}$ and $f^\star$ represent the best observation of the objective function found in each iteration and its true minimum value, respectively.
We set the number of initial observations and the number of BO iterations at $N=5d$ and $K=50$ for the 2d Ackley function, and $N=10d$ and $K=200$ for the 6d Rosenbrock function.

\begin{figure}[t]
	\centering
	\begin{subfigure}[c]{0.49\textwidth}
		\centering
		\includegraphics[width=\hsize]{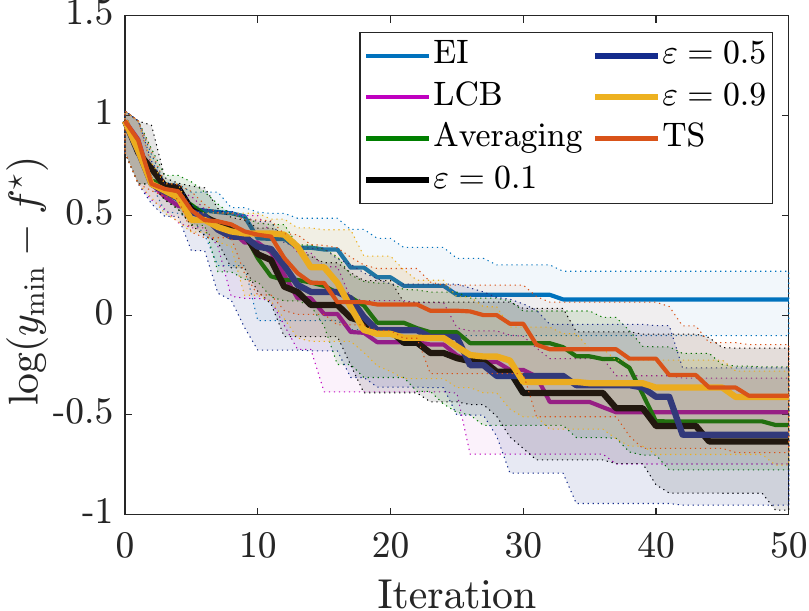}
		\caption{}
		\label{fig:TestFunctions_a}
	\end{subfigure}
	\begin{subfigure}[c]{0.49\textwidth}
		\centering
		\includegraphics[width=\hsize]{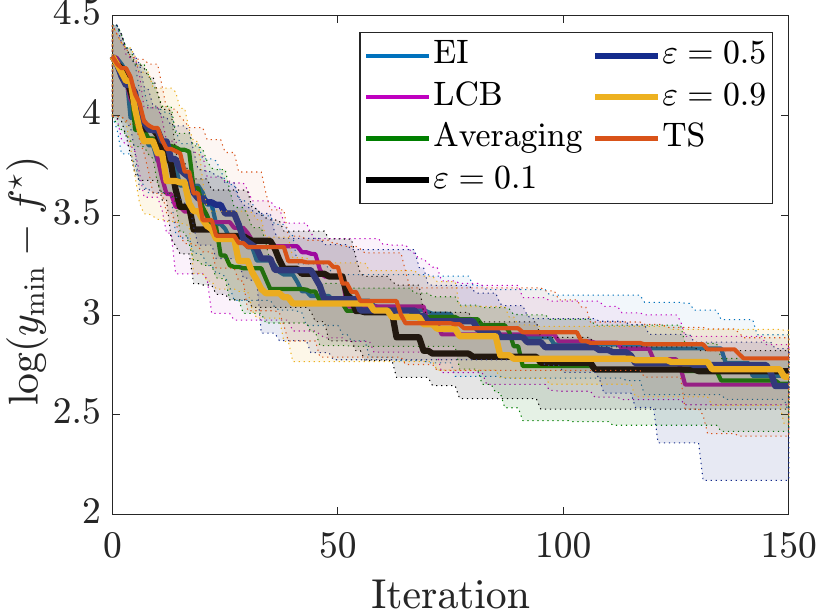}
		\caption{}
		\label{fig:TestFunctions_b}
	\end{subfigure}

	\begin{subfigure}[c]{0.49\textwidth}
		\centering
		\includegraphics[width=\hsize]{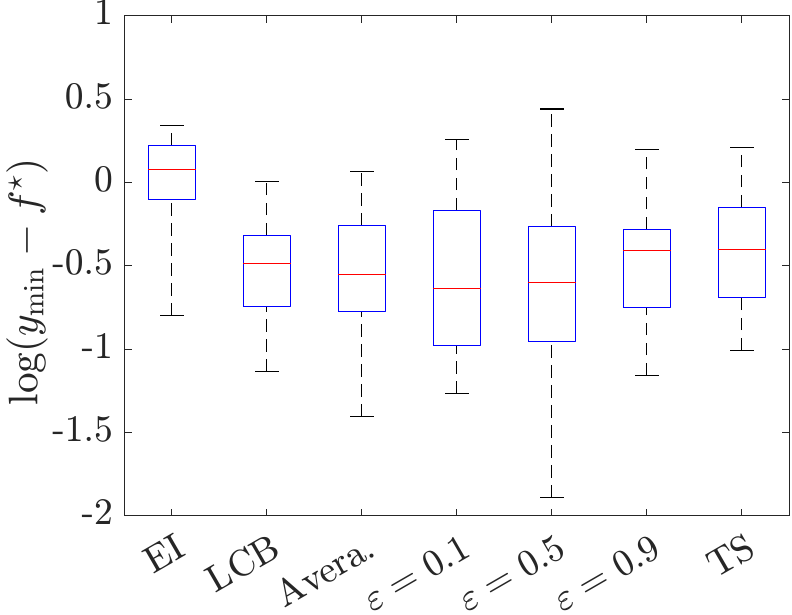}
		\caption{}
		\label{fig:TestFunctions_c}
	\end{subfigure}
	\begin{subfigure}[c]{0.49\textwidth}
		\centering
		\includegraphics[width=\hsize]{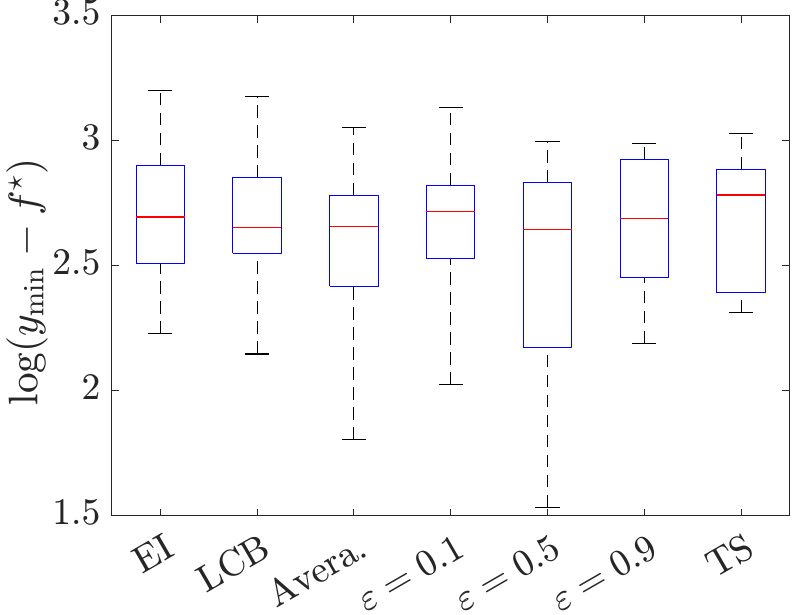}
		\caption{}
		\label{fig:TestFunctions_d}
	\end{subfigure}
	
	\caption{Performance of EI, LCB, averaging TS, generic TS, and $\varepsilon$-greedy TS methods for the 2d Ackley and 6d Rosenbrock functions. Optimization histories for (a) the 2d Ackley function and (b) the 6d Rosenbrock function. Medians and interquartile ranges of final solutions from 100 runs of each BO method for (c) the 2d Ackley function and (d) the 6d Rosenbrock function.} 
	\label{fig:TestFunctions}
\end{figure}

\Cref{fig:TestFunctions_a,fig:TestFunctions_b} show the medians and interquartile ranges obtained from 100 runs of each BO method for the 2d Ackley and 6d Rosenbrock functions, respectively.
\Cref{fig:TestFunctions_c,fig:TestFunctions_d} show the medians and interquartile ranges of final solutions.
In both cases, the optimization results from $\varepsilon$-greedy TS for an appropriate $\varepsilon$ are better than those from one of its two extremes and competitive with the results from the other.
In addition, $\varepsilon$-greedy TS for $\varepsilon = 0.5$ can provide the best objective values among those from the considered BO methods.

\begin{figure}[t]
	\centering
	\begin{subfigure}[c]{0.49\textwidth}
		\centering
		\includegraphics[width=\hsize]{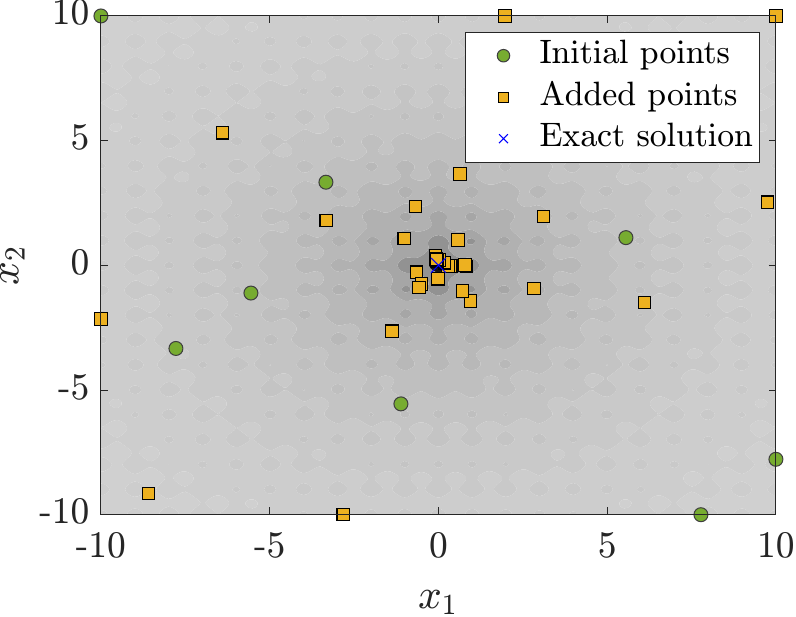}
		\caption{Averaging TS, $\varepsilon = 0$}
		\label{fig:AddedPoints_a}
	\end{subfigure}
	\begin{subfigure}[c]{0.49\textwidth}
		\centering
		\includegraphics[width=\hsize]{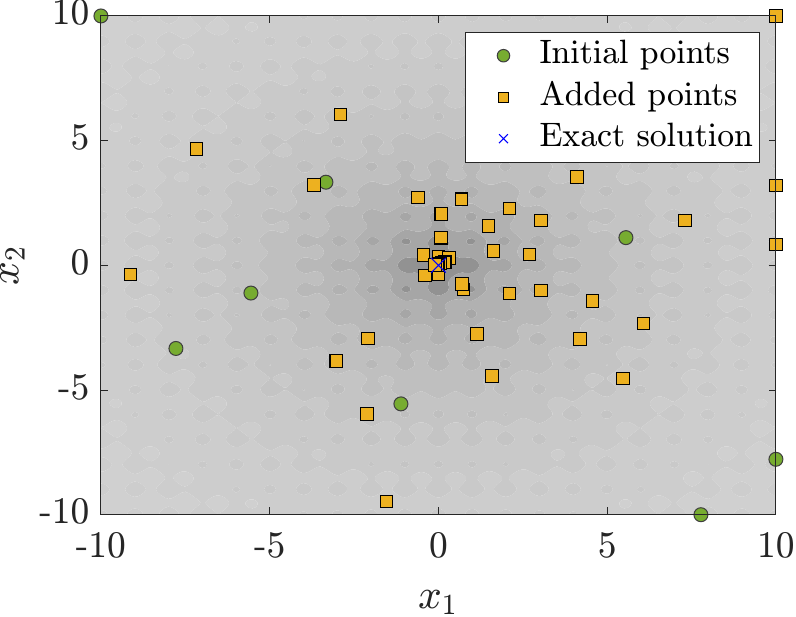}
		\caption{$\varepsilon$-greedy TS, $\varepsilon = 0.1$}
		\label{fig:AddedPoints_b}
	\end{subfigure}
	
	\begin{subfigure}[c]{0.49\textwidth}
		\centering
		\includegraphics[width=\hsize]{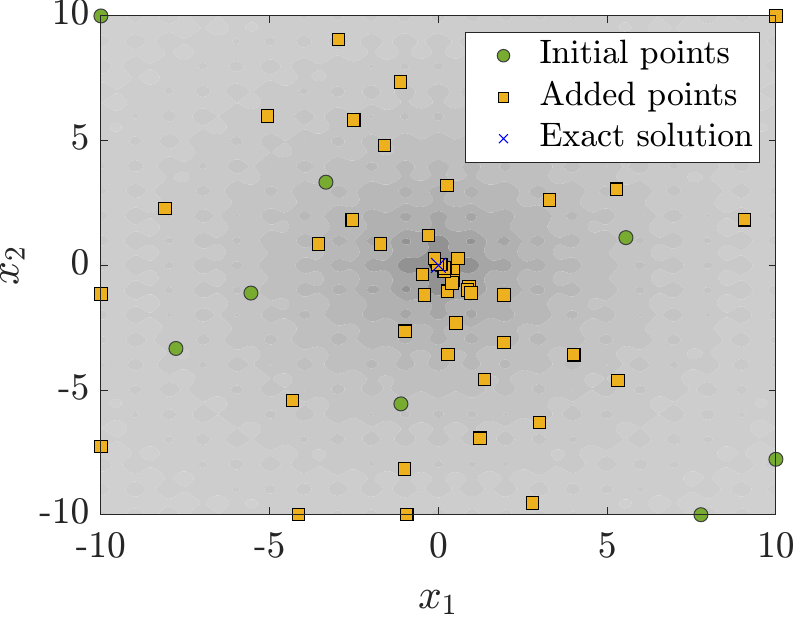}
		\caption{$\varepsilon$-greedy TS, $\varepsilon = 0.9$}
		\label{fig:AddedPoints_c}
	\end{subfigure}
	\begin{subfigure}[c]{0.49\textwidth}
		\centering
		\includegraphics[width=\hsize]{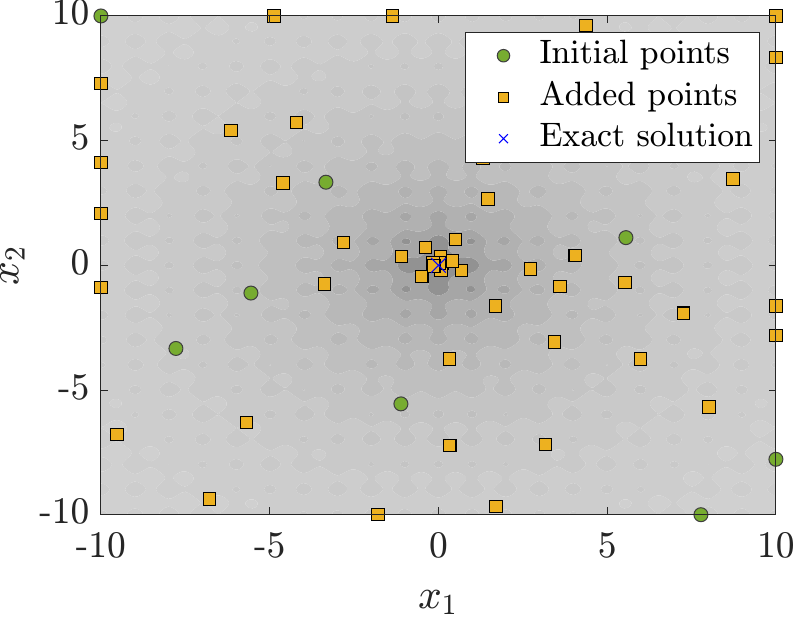}
		\caption{Generic TS, $\varepsilon = 1$}
		\label{fig:AddedPoints_d}
	\end{subfigure}
	\caption{Initial and added points for 2d Ackley function with $N_\text{s} = 50$ and different values of $\varepsilon$.} 
	\label{fig:AddedPoints}
\end{figure}

While the optimization results in \cref{fig:TestFunctions} suggest that it is safe to set $\varepsilon=0.5$ to balance exploitation and exploration, an optimal $\varepsilon$ may depend on the nature of the problem of interest.
A problem that requires more exploitation for its optimal solution (e.g., the 2d Ackley function) may benefit from a smaller $\varepsilon$.
A problem with more exploration (e.g., the 6d Rosenbrock function) might be better addressed with a larger $\varepsilon$.
Yet determining whether a black-box objective function leans toward exploitation or exploration for its extrema is often elusive.
This, ironically, underscores the rationale behind employing $\varepsilon$-greedy TS to prevent our search from overly favoring either exploitation or exploration.
In $\varepsilon$-TS for multi-armed bandits, $\varepsilon$ can be set as $\varepsilon = 1/N_{a}$ for achieving simultaneous minimax and asymptotic optimality \cite{Jin2023}, where $N_{a}$ is the number of arms to select. 
This, however, does not apply to BO because the number of new solution points, which are similar to arms in multi-armed bandits, is uncountable.
It is also worth noting that any endeavors to find an optimal value of $\varepsilon$ for a particular problem should correspond to a specific accuracy level of the initial GP model because lowering the model fidelity always encourages exploration, which is irrespective of the selection of $\varepsilon$.

As discussed in \cref{sec:introduction}, increasing $\varepsilon$ value results in more exploration of the $\varepsilon$-greedy strategy, and this holds true for $\varepsilon$-greedy TS.
An increase in $\varepsilon$ encourages the algorithm to explore more unseen regions of the input variable space, which is confirmed by observing how the algorithm, with varying $\varepsilon$ values, selects new solution points for the Ackley function, as shown in \cref{fig:AddedPoints}.

\begin{figure}[t]
	\centering
	\begin{subfigure}[c]{0.49\textwidth}
		\centering
		\includegraphics[width=\hsize]{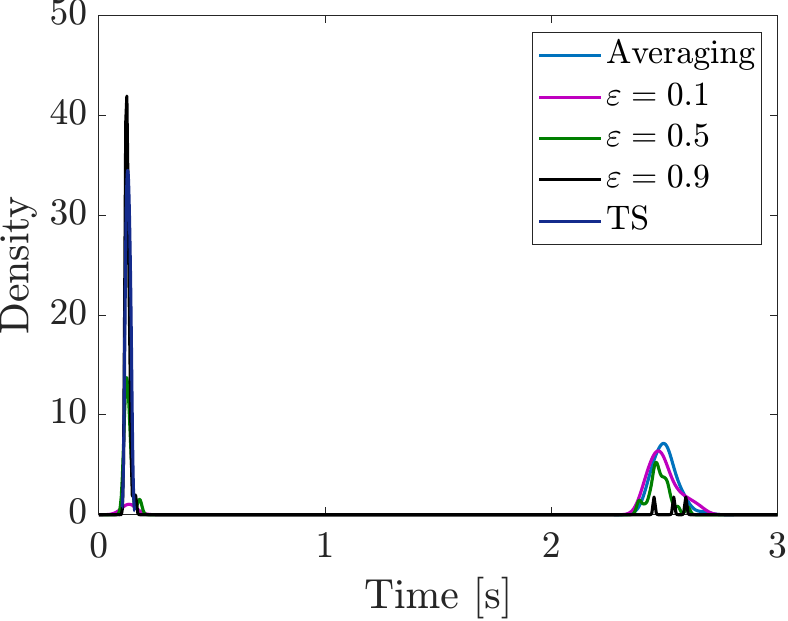}
		\caption{2d Ackley}
		\label{fig:RunTime_a}
	\end{subfigure}
	\begin{subfigure}[c]{0.49\textwidth}
		\centering
		\includegraphics[width=\hsize]{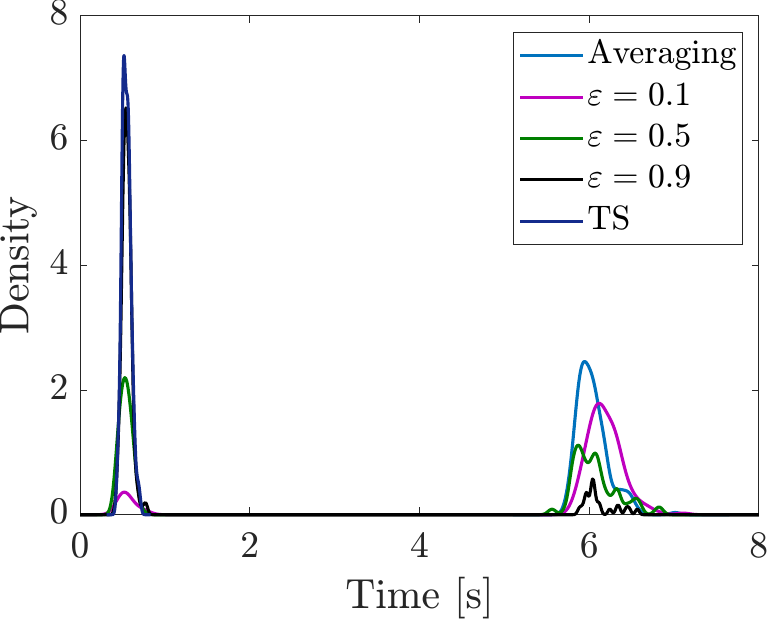}
		\caption{6d Rosenbrock}
		\label{fig:RunTime_b}
	\end{subfigure}
	\caption{Approximate distributions of runtime for selecting a new solution point with different $\varepsilon$ values and $N_\text{s} = 50$.} 
	\label{fig:RunTime}
\end{figure}

We also investigate the effect of varying $\varepsilon$ on the runtime for selecting a new solution point.
\Cref{fig:RunTime} shows approximate distributions of runtime for selecting a new solution point by the sample-average TS with $N_\text{s}=50$, $\varepsilon$-greedy TS with different $\varepsilon$ values and $N_\text{s}=50$, and the generic TS for the 2d Ackley and 6d Rosenbrock functions.
There are two distinct clusters of the runtime values for $\varepsilon$-greedy TS.
One cluster aligns with the runtime of the generic TS and the other corresponds to the runtime of the sample-average TS.
In addition,
the average runtimes for each BO trial of the sample-average TS,
$\varepsilon$-greedy TS with $\varepsilon=$ 0.1, 0.5, 0.9, and the generic TS
are 124, 102, 64, 18, and 7 s for the 2d Ackley function, respectively.
Those for the 6d Rosenbrock functions are 1377, 1188, 809, 229, and 111 s, respectively.
We see that increasing $\varepsilon$ reduces the runtime of each BO trial.
This can be attributed to the fact that a larger $\varepsilon$ requires more exploration of calling the generic TS, which is cheaper than the sample-average TS.

\subsection{Cantilever Beam}
\label{sec:example-beam}

\begin{figure*}[t]
	\centering
	\includegraphics[scale=0.55]{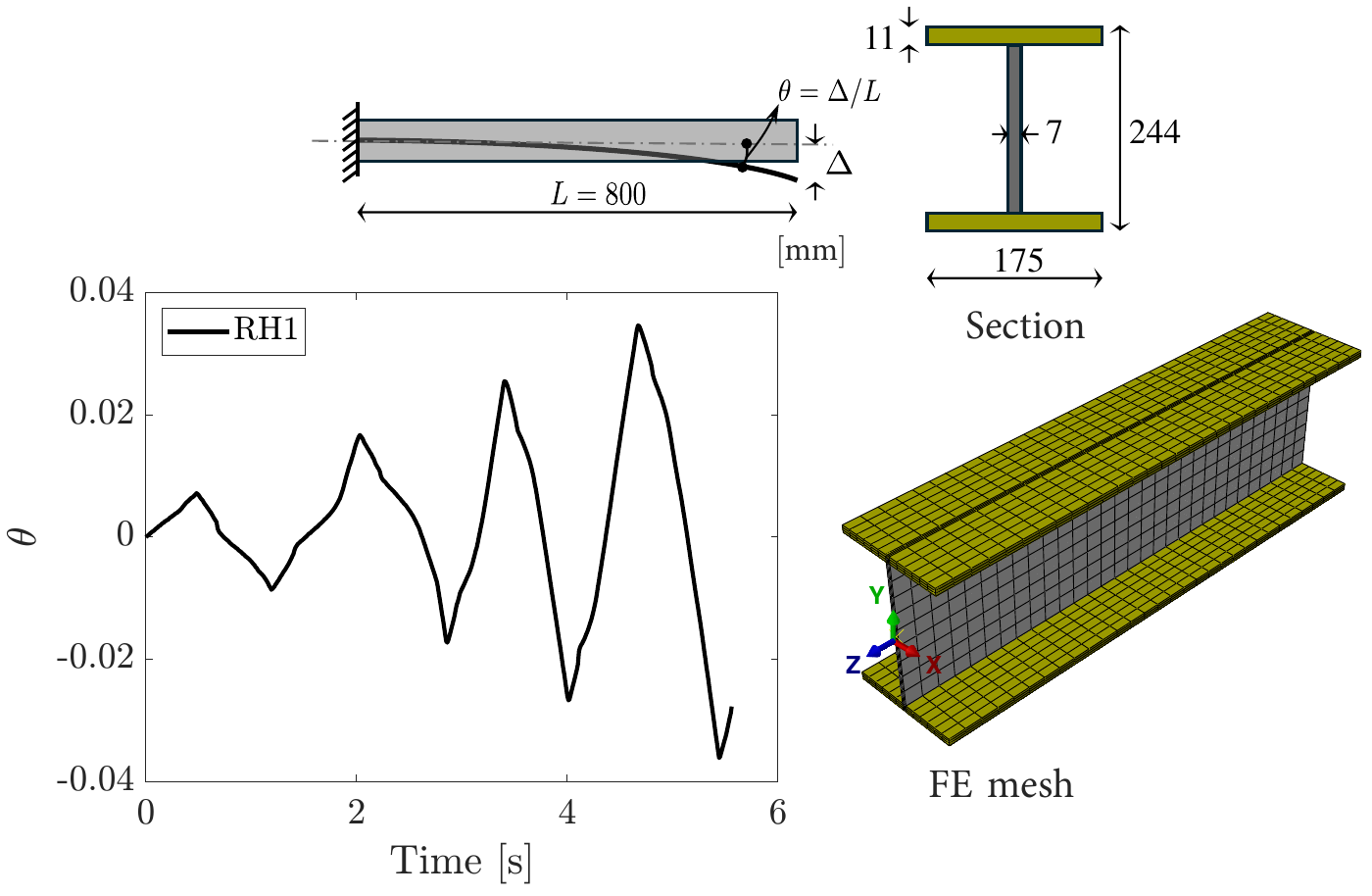}
	\caption{The cantilever beam, its finite-element (FE) mesh, and loading history RH1 for cyclic test~\cite{Do2022}.}
	\label{fig:beamModel}
\end{figure*}

In this example, we use $\varepsilon$-greedy TS to identify the constitutive parameters for simulating the behavior of a steel cantilever beam under a static cyclic loading history RH1, as shown in \cref{fig:beamModel}. The web and flange of the beam have different material parameters. In the experiment conducted by Yamada and Jiao~\cite{Yamada2016}, forced vertical displacement was applied at the free end of the beam. The defection angle $\theta$ was defined as the ratio of the vertical tip displacement $\Delta$ mm to the beam length $ L = 800$ mm, i.e., $\theta = \Delta/L$.
Values of moment reaction $M$ about $x$-axis at the beam support were measured.

Let ${\bf x}$ represent the vector of parameters underlying a constitutive model for the cyclic behavior of the beam, for which the detailed description is given in \cref{sec:example-beam-materialmodel}. Let $M_{t}^\text{s}$ represent the simulated value of the moment reaction at the $t$th time step of a discretized loading history of RH1 consisting of $T$ steps, and $M_{t}^\text{m}$ the corresponding measured value.
We formulate the following misfit function $f(\textbf{x})$ for finding an optimal set of ${\bf x}$~\citep{Ohsaki2016}:
\begin{equation}\label{eqn13}
	f({\bf x}) = \sqrt{\frac{1}{T}\displaystyle\sum_{t=1}^{T}\left[M_{t}^\text{s}({\bf x})-M_{t}^\text{m}\right]^2}.
\end{equation}

To evaluate $f({\bf x})$, we model the beam using Abaqus 2022.
We reduce the sensitivity of simulation estimates to the finite-element mesh density by generating a fine mesh of 4960 nodes and 3510 linear hexahedral elements of type C3D8, see \cref{fig:beamModel}.
The maximum increment size for each loading history is set as 0.01 s.

\subsubsection{Nonlinear Combined Isotropic/Kinematic Hardening}
\label{sec:example-beam-materialmodel}

Following \cite{Do2022}, we capture the cyclic elastoplastic behavior of the beam using the nonlinear combined isotropic/kinematic hardening model~\cite{Lemaitre1994}. This model combines the properties of both isotropic and nonlinear kinematic hardening to describe the relationship between strain and stress states at each time instant.
This relationship is established based on the material status that, either elastic or plastic, is detected by the following von Mises yield condition:
\begin{equation}\label{eqn14}
	F=\|\boldsymbol{\xi}\|-\sqrt{\frac{2}{3}}\sigma_{\text{y}}\leq 0,
\end{equation}
where $\boldsymbol{\xi}=\text{dev}[\boldsymbol{\sigma}]-\text{dev}[\boldsymbol{\alpha}]$ is the shifted-stress tensor, $\text{dev}[\cdot]$ the deviatoric part of $[\cdot]$, $\|\cdot\|$ the $L^2$-norm of the tensor, $\boldsymbol{\sigma}$ the point stress tensor, $\boldsymbol{\alpha}$ the back-stress tensor, and $\sigma_{\text{y}}$ the yield stress.

The isotropic hardening increases the size of the yield surface $F=0$ during the evolution of plastic deformation, while fixing the shape and location of this surface.
In this case, $\boldsymbol{\xi}$ in \cref{eqn14} does not involve the back-stress tensor $\boldsymbol{\alpha}$, resulting in an isotropic yield surface of the stress.
Since structural steels exhibit a saturation point of the stress at large deformation, the isotropic hardening for them models the increment of yield surface size using the following monotonically increasing nonlinear function~\cite{Voce1948}:
\begin{equation}\label{eqn15}
	\sigma_{\text{y}}=\sigma_{\text{y},0}+Q_\infty[1-\exp(-b\epsilon_{\text{eq}}^{\text{p}})],
\end{equation}
\noindent
where $\sigma_{\text{y},0}$ is the initial yield stress, $Q_\infty$ the difference of the stress saturation and $\sigma_{\text{y},0}$, $b$ the isotropic saturation rate, and $\epsilon_{\text{eq}}^{\text{p}}$ the current equivalent plastic strain determined based on its previous state and the associated plastic strain rate $\dot{\epsilon}_{\text{eq}}^\text{p}$.

Unlike the isotropic hardening, the kinematic hardening does not change the size and shape of the yield surface during the evolution of plastic deformation.
Instead, it relocates the center of the yield surface by a rigid translation in the evolution direction of the plastic strain.
This allows the model to capture the Bauschinger effect.
In pure kinematic hardening, the yield function in \cref{eqn14} has $\sigma_\text{y}$ fixed at $\sigma_{\text{y},0}$ and the back-stress tensor $\boldsymbol{\alpha}$ defined as the superposition of $n_{\text{k}}$ back-stress components, such that~\cite{Chaboche1983}
\begin{equation}\label{eqn16}
	\boldsymbol{\alpha}=\displaystyle\sum_{k=1}^{n_{\text{k}}} \boldsymbol{\alpha}_k.
\end{equation}
Here the evolution of the $k$th component $\boldsymbol{\alpha}_k$ is described by a nonlinear kinematic hardening rule, as~\cite{Armstrong1966}
\begin{equation}\label{eqn17}
	\dot{\boldsymbol{\alpha}}_k=\sqrt{\frac{2}{3}}C_k\dot{\epsilon}_{\text{eq}}^{\text{p}}\textbf{n}-\gamma_k\dot{\epsilon}_{\text{eq}}^{\text{p}}\boldsymbol{\alpha}_k,
\end{equation}
where $\textbf{n}= \boldsymbol{\xi}/\|\boldsymbol{\xi}\|$, and $C_k$ and $\gamma_k$ the translation and relaxation rates of the back-stress component $k$, respectively.

The nonlinear combined isotropic/kinematic hardening model makes use of \cref{eqn15,eqn16,eqn17} for checking the material status at each time instant. If we use one back-stress component for this model, its underlying parameters read $\textbf{x}=[E,\nu,Q_\infty,b,\sigma_{\text{y},0},C_1,\gamma_1]^\intercal$, where Young's modulus $E$ and Poisson's ratio $\nu$ are incorporated in the isotropic elastic tensor.

\begin{table}[tb]
	\caption{Material parameter intervals for the cantilever beam.}
	\label{table2}
	\centering
	\begin{tabular}{llcc}
		\hline\noalign{\smallskip}
		Component & Parameter & Lower bound & Upper bound\\
		\hline\noalign{\smallskip}
		\multirow{5}{*}{\parbox[c]{.08\linewidth}{Web}} & $E$ [GPa] & $175.05$   &$-$  \\
		& $\nu$ & $0.3$   &$-$  \\
		& $\sigma_{\text{y},0}$ [MPa] & $300$ & $340$ \\
		& $Q_\infty$ [MPa] & $10$ & $100$ \\
		& $b$ & $5$ &$30$ \\
		& $C_1$ [MPa] & $2 \times 10^3$ &$10^4$ \\
		& $\gamma_1$ & $10$ &$200$ \\
		\hline\noalign{\smallskip}
		\multirow{5}{*}{\parbox[c]{.08\linewidth}{Flange}} & $E$ [GPa] & $175.05$   &$-$  \\
		& $\nu$ & $0.3$   &$-$  \\
		& $\sigma_{\text{y},0}$ [MPa] & $270$ & $300$ \\
		& $Q_\infty$ [MPa] & $10$ & $100$ \\
		& $b$ & $5$ &$30$ \\
		& $C_1$ [MPa] & $2 \times 10^3$ &$10^4$ \\
		& $\gamma_1$ & $10$ &$200$ \\
		\hline\noalign{\smallskip}
	\end{tabular}
\end{table}

\subsubsection{Constitutive Parameters by Bayesian Optimization}
\label{sec:example-beam-results}

During the minimization of misfit function $f(\mathbf{x})$, we keep Young's modulus and Poisson’s ratio for the web and flange of the beam as constants, set at $E=175.05$ GPa and $\nu=0.3$~\cite{Yamada2016}.
As a result, we identify ten parameters for the beam of which five parameters are associated with the web and the remaining five with the flange.
The permissible range for each parameter is taken from~\cite{Do2022} and listed in~\cref{table2}.
To initiate BO, we generate ten distinct initial datasets, half containing 50 data points each and the remainder 100 data points each.
With a total of $100$ BO iterations planned, we execute the optimization process using the Carya Cluster housed at the University of Houston.
This cluster hosts 9984 Intel CPU cores and 327680 Nvidia GPU cores integrated within 188 compute and 20 GPU nodes.
This infrastructure allows us to complete a costly beam simulation in three minutes.

\begin{figure}[t]
	\centering
	\begin{subfigure}[c]{0.49\textwidth}
		\centering
		\includegraphics[width=\hsize]{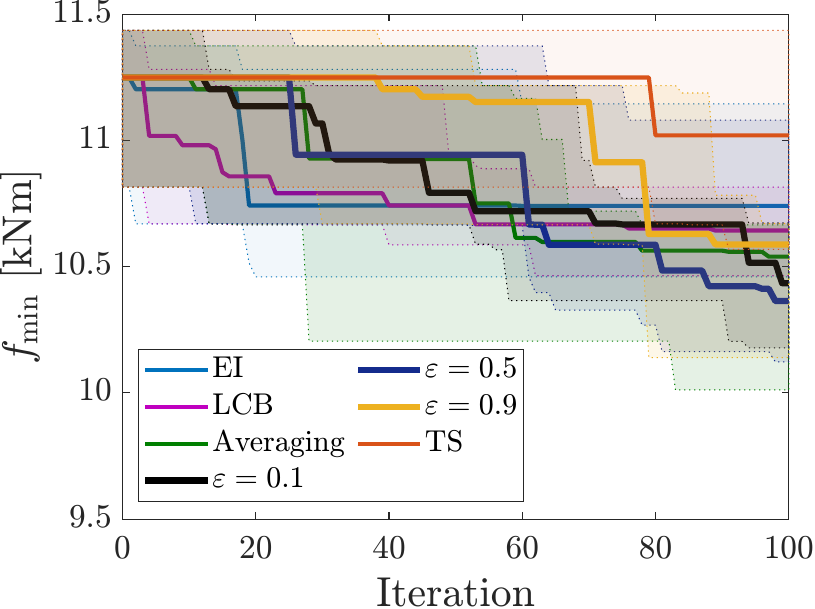}
		\caption{}
		\label{fig:BeamHistory_a}
	\end{subfigure}
	\begin{subfigure}[c]{0.49\textwidth}
		\centering
		\includegraphics[width=\hsize]{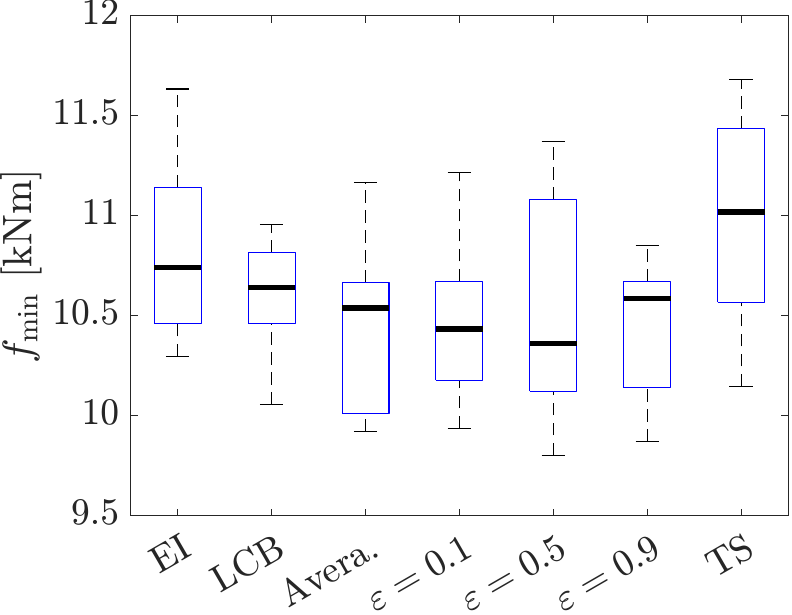}
		\caption{}
		\label{fig:BeamHistory_b}
	\end{subfigure}
	\caption{Performance of EI, LCB, averaging TS, generic TS, and $\varepsilon$-greedy TS methods for the cantilever beam. (a) Optimization histories; (b) Medians and interquartile ranges of final solutions from ten runs of each BO method.} 
	\label{fig:BeamHistory}
\end{figure}

\Cref{fig:BeamHistory} shows the optimization histories and variations in the final solution obtained from different BO methods for the cantilever beam, where $f_{\min}$ represents the best observation of $f({\bf x})$ found in each BO iteration.
The identification results obtained from $\varepsilon-$greedy TS with $\varepsilon = 0.1$ and $0.5$
are comparable to or better than that from the sample-average TS,
and are better than that from the generic TS.
More notably, $\varepsilon-$greedy TS for $\varepsilon = 0.5$ provides the best set of identified parameters.

\begin{figure}[t]
	\centering
	\begin{subfigure}[c]{0.32\textwidth}
		\centering
		\includegraphics[width=\hsize]{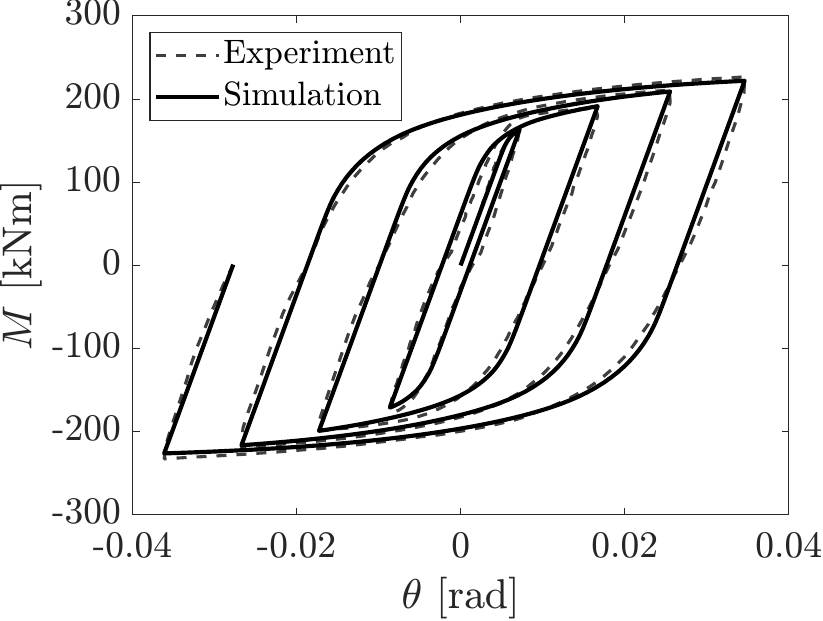}
		\caption{}
		\label{fig:Beamprediction_a}
	\end{subfigure}
	\begin{subfigure}[c]{0.32\textwidth}
		\centering
		\includegraphics[width=\hsize]{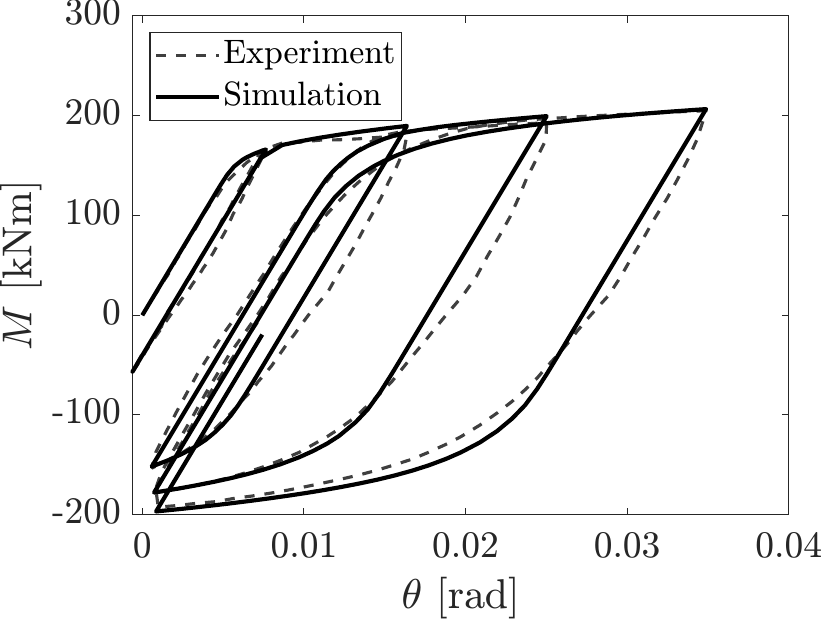}
		\caption{}
		\label{fig:Beamprediction_b}
	\end{subfigure}
	\begin{subfigure}[c]{0.32\textwidth}
		\centering
		\includegraphics[width=\hsize]{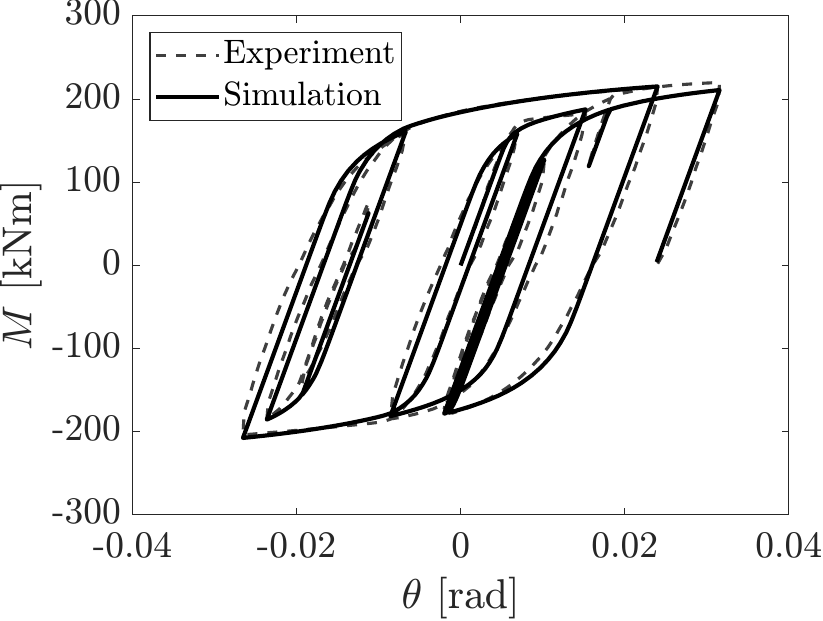}
		\caption{}
		\label{fig:Beamprediction_c}
	\end{subfigure}
	\caption{Measured and simulated $M$--$\theta$ curves of the cantilever beam under different cyclic loading histories. The simulated $M$--$\theta$ curves correspond to the best parameter set identified from $\varepsilon$-TS and the experimental results from RH1. (a) RH1; (b) RH2; (c) RH3. Here experimental data obtained from RH2 and RH3 are unseen by the identification process.} 
	\label{fig:Beamprediction}
\end{figure}

We use the best set of parameters identified from $\varepsilon$-greedy TS to predict the $M$--$\theta$ curves of the beam under RH1 and other two cyclic loading histories RH2 and RH3.
The experimental data associated with RH2 and RH3 were not fed to the parameter identification process.
The agreement observed between the measured and simulated $M$--$\theta$ curves for each loading history, as shown in \cref{fig:Beamprediction}, confirms the reliable prediction performance achieved with the identified parameters.

\section{Conclusions}
\label{sec:conclusions}

We introduced $\varepsilon$-greedy Thompson sampling (TS) to optimizing costly simulation-based objective functions.
The method addresses the classic exploitation--exploration dilemma by randomly switching between its two extremes, namely the generic TS and the sample-average TS.
Our empirical findings reveal that $\varepsilon$-greedy TS with an appropriate $\varepsilon$ value
robustly achieves the better performance of its two extremes across the benchmark problems,
and can provide the best result in a practical engineering application.

While several $\varepsilon$-greedy algorithms and the generic TS are guaranteed to converge eventually~\cite{DeAth2021,Garnett2023}, we look forward to a theoretical analysis to elucidate the convergence properties of $\varepsilon$-greedy TS.
The regret bound analysis of $\varepsilon$-TS \cite{Jin2023} for multi-armed bandits can serve as a starting point for this endeavor.
We are also interested in exploring extensions of $\varepsilon$-greedy TS to high-dimensional settings where good approximations of the GP posteriors and efficient optimization of TS acquisition functions in high dimensions are of interest. For this, subspace-based approaches~\cite{Nayebi2019,ZhangRD2022gps} or trust-region methods~\cite{Eriksson2019} present viable strategies.
Additionally, the application of $\varepsilon$-greedy TS to non-Gaussian likelihood settings and the evolution of $\varepsilon$ during the optimization process are still open problems. The former may be addressed using a Markov chain Monte Carlo algorithm (see e.g., \cite{Mazumdar2020,ZhengH2024}) to sample a large set of function values or to sample the weight values $\boldsymbol{\beta}$ in \cref{eqn8} from a non-Gaussian posterior.

\section*{Acknowledgments}

We thank the University of Houston for providing startup fund to support this research.
We thank Prof. Satoshi Yamada at the University of Tokyo and Prof. Makoto Ohsaki at Kyoto University for providing experimental results of the cantilever beam used in \cref{sec:example-beam}.

\clearpage
\bibliographystyle{elsarticle-num-names}
\bibliography{GreedyBO} 

\end{document}